\documentclass[conference]{IEEEtran}
\IEEEoverridecommandlockouts
\usepackage{algorithmic}
\usepackage{array}
\usepackage[caption=false,font=normalsize,labelfont=sf,textfont=sf]{subfig}
\usepackage{textcomp}
\usepackage{stfloats}
\usepackage{url}
\usepackage{verbatim}
\usepackage{graphicx}
\usepackage{rotating}  

\usepackage{algorithm}          
\usepackage{algorithmic}        

\usepackage{pifont}

\usepackage{multirow,bbding}
\usepackage{amsmath,amssymb,amsfonts,bm}
\usepackage{threeparttable}

\usepackage{booktabs}
\usepackage{bbding}

\usepackage{tcolorbox}
\usepackage{mdframed}

\usepackage{wasysym}

\usepackage{cite}

\usepackage{tablefootnote}

\usepackage{longtable}
\usepackage{newclude}
\def\BibTeX{{\rm B\kern-.05em{\sc i\kern-.025em b}\kern-.08em
    T\kern-.1667em\lower.7ex\hbox{E}\kern-.125emX}}
\begin{document}

\title{M²S²L: Mamba-based Multi-Scale Spatial-temporal Learning for Video Anomaly Detection
}

\author{
    \IEEEauthorblockN{
        Yang Liu\textsuperscript{1},
        Boan Chen\textsuperscript{2},
        Xiaoguang Zhu\textsuperscript{3},
        Jing Liu\textsuperscript{4,5*},
        Peng Sun\textsuperscript{5*}\thanks{$^*$Corresponding authors. This work was supported by the Professional Discretionary Fund under Grant No. 26AKUG0088.}, 
        Wei Zhou\textsuperscript{6}
    }

    \IEEEauthorblockA{
        \textsuperscript{1}Tongji University \quad \textsuperscript{2}SJTU \quad \textsuperscript{3}UC Davis \quad \textsuperscript{4}UBC \quad \textsuperscript{5}Duke Kunshan University \quad \textsuperscript{6}Cardiff University
        \\
\{yang\_liu, jingliu\}@ieee.org, bchen4@sjtu.edu.cn, xgzhu@ucdavis.edu, peng.sun568@duke.edu, zhouw26@cardiff.ac.uk
    }
}

\maketitle

\begin{abstract}
Video anomaly detection (VAD) is an essential task in the image processing community with prospects in video surveillance, which faces fundamental challenges in balancing detection accuracy with computational efficiency. As video content becomes increasingly complex with diverse behavioral patterns and contextual scenarios, traditional VAD approaches struggle to provide robust assessment for modern surveillance systems. Existing methods either lack comprehensive spatial-temporal modeling or require excessive computational resources for real-time applications. In this regard, we present a Mamba-based multi-scale spatial-temporal learning (M²S²L) framework in this paper. The proposed method employs hierarchical spatial encoders operating at multiple granularities and multi-temporal encoders capturing motion dynamics across different time scales. We also introduce a feature decomposition mechanism to enable task-specific optimization for appearance and motion reconstruction, facilitating more nuanced behavioral modeling and quality-aware anomaly assessment. Experiments on three benchmark datasets demonstrate that M²S²L framework achieves 98.5\%, 92.1\%, and 77.9\% frame-level AUCs on UCSD Ped2, CUHK Avenue, and ShanghaiTech respectively, while maintaining efficiency with 20.1G FLOPs and 45 FPS inference speed, making it suitable for practical surveillance deployment.
\end{abstract}


\section{Introduction}

Video Anomaly Detection (VAD) has emerged as a pivotal technology for intelligent surveillance systems, aiming to automatically identify irregular events or behaviors that deviate from established normal patterns in video sequences~\cite{liu2025networking,wu2024domain}. As modern surveillance ecosystems evolve to encompass diverse behavioral modeling and context-aware assessment requirements \cite{liu2025edge,liu2025domain}, traditional anomaly detection approaches face new challenges in understanding complex spatio-temporal patterns and user interaction scenarios. With the exponential growth of surveillance infrastructure across urban environments, transportation networks, and industrial facilities \cite{10646471,10409541}, the demand for automated monitoring solutions that can provide real-time quality assurance and perceptual understanding has intensified dramatically, making manual oversight increasingly impractical and resource-intensive~\cite{akdag2024teg,liu2023stochastic,zhou2025video,liu2024generalized}.

Current unsupervised VAD methodologies can be broadly categorized into two primary paradigms: \textit{Single-stream normality learning} \cite{liu2025anomaly} and \textit{Multi-stream normality learning} \cite{liu2025privacy}. Single-stream approaches~\cite{zhao2025rethinking,liu2018future,liu2023learning,cheng2024denoising} typically process entire video sequences through unified architectures, learning compact representations of normal patterns. While computationally efficient, these methods often struggle to adequately model the inherent differences between visual appearance and temporal motion characteristics, leading to suboptimal detection performance in complex scenarios. Conversely, multi-stream approaches~\cite{chang2022video,liu2024memory,rodrigues2021multi,liu2022appearance} explicitly decompose video understanding into separate appearance and motion processing pathways, enabling more nuanced modeling of distinct visual modalities~\cite{liu2025crcl}. However, this enhanced representational capability comes at the cost of significantly increased computational overhead, often requiring multiple CNN or Transformer-based encoders that scale quadratically with sequence length~\cite{huang2022multimodal,liu2023learning}.

Recently, state space models, particularly Mamba~\cite{gu2023mamba}, have demonstrated remarkable potential for efficient sequence modeling with linear computational complexity. Initial explorations in VAD, including STNMamba~\cite{li2024stnmamba} and VADMamba~\cite{lyu2025vadmamba}, have attempted to leverage these advantages for balancing detection accuracy and computational efficiency. However, existing Mamba-based VAD methods face critical limitations that hinder their practical deployment. Specifically, these approaches typically employ homogeneous encoder architectures that process both appearance and motion information using identical structures, failing to capture the distinct characteristics of visual appearance and temporal motion patterns. Moreover, they often lack effective mechanisms for handling the multi-scale nature of anomalous events, which can manifest across different spatial granularities and temporal durations in real-world surveillance scenarios.

\begin{figure}[t]
\centering
\includegraphics[width=0.48\textwidth]{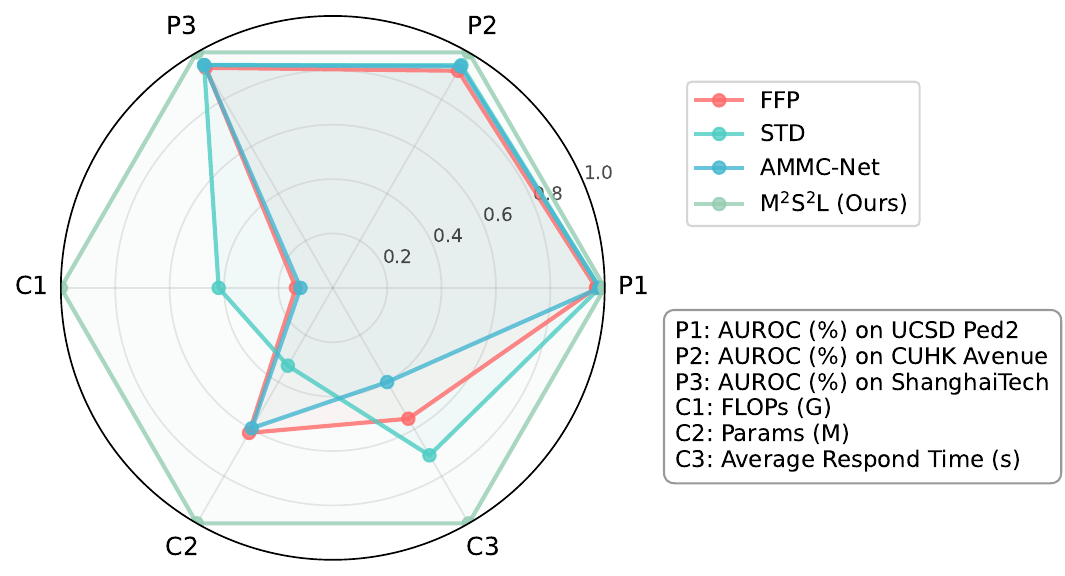}
\caption{Evaluation of different VAD approaches on detection accuracy and efficiency measures. The outer vertices correspond to AUC on P1: UCSD Ped2, P2: CUHK Avenue, and P3: ShanghaiTech, while inner vertices represent efficiency indicators (C1: FLOPs, C2: Model Parameters, C3: Response Time). Values are scaled proportionally with optimal performance set to 1.0. 
}
\label{fig:radar_comparison}
\vspace{-5pt}
\end{figure}

\begin{figure*}[t]
\centering
\includegraphics[width=0.98\textwidth]{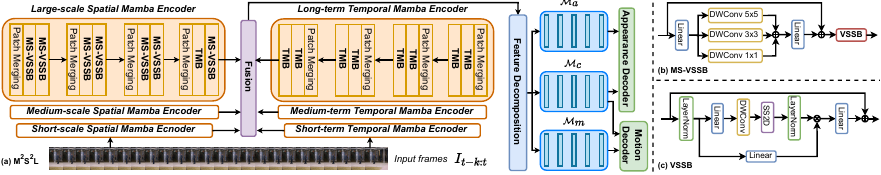}
\caption{Architecture overview of (a) M²S²L framework, (b) Multi-Scale Visual State Space Block (MS-VSSB), and (c) Standard Visual State Space Block (VSSB). Multi-scale spatial and temporal Mamba encoders process input sequences at different granularities through parallel processing streams. 
}
\label{fig:overall_architecture}
\vspace{-15pt}
\end{figure*}

To address these limitations, we propose the Mamba-based multi-scale spatial-temporal learning (M²S²L) for VAD in this paper. Our approach introduces specialized dual-stream Mamba encoders that separately model appearance and motion normality patterns, enabling task-specific feature learning while preserving computational efficiency. Additionally, we develop hierarchical multi-scale processing mechanisms that capture spatial-temporal patterns across different granularities, from fine-grained local anomalies to coarse-grained scene-level irregularities. Through these architectural designs, M²S²L achieves detection performance comparable to existing methods while maintaining computational efficiency suitable for real-time video surveillance systems (20.1G FLOPs, 14.9M parameters, and 45 FPS inference speed). The radar chart comparison in Fig.~\ref{fig:radar_comparison} demonstrates that M²S²L achieves superior balance across all evaluation metrics, establishing new benchmarks for the trade-off between detection accuracy and computational feasibility in practical surveillance applications.

The main contributions are summarized as follows:
\begin{itemize}
    \item We introduce a dual encoding and dual decoding architecture that separately learns appearance and motion normality patterns, enabling specialized processing of distinct visual modalities while maintaining computational efficiency through Mamba-based state space modeling.
    \item We develop multi-scale spatial-temporal normality learning mechanisms that capture anomalous patterns across different spatial granularities and temporal durations, addressing the scale-variant nature of anomalous events in surveillance scenarios.
    \item We demonstrate that our approach achieves competitive detection performance with significantly reduced computational overhead, making it suitable for deployment in real-time video surveillance systems.
\end{itemize}

\section{Methodology}

\subsection{Overview}

The proposed M²S²L framework adopts a dual-stream architecture designed to efficiently model both spatial appearance and temporal motion normality patterns. As illustrated in Fig.~\ref{fig:overall_architecture}, the system consists of three core components: \textbf{(1)} multi-scale spatial normality learning for hierarchical appearance pattern modeling, \textbf{(2)} multi-scale temporal normality learning for motion dynamics across different time scales, and \textbf{(3)} feature decomposition with specialized decoding for task-specific reconstruction. Given input video clips $\mathcal{V}_{t-k:t} = \{V_{t-k}, V_{t-k+1}, \ldots, V_t\} \in \mathbb{R}^{k \times H \times W \times 3}$ where $k$ denotes the temporal window length, the M²S²L framework simultaneously learns appearance and motion normality through parallel processing streams. During training, the framework exclusively observes normal sequences to establish baseline patterns. At inference time, anomalous events are identified through reconstruction discrepancies in both modalities.

\subsection{Multi-Scale Spatial Normality Learning}

The spatial normality learning component captures appearance patterns across multiple spatial granularities to handle anomalies that manifest at different scales, from localized texture changes to scene-level variations. The hierarchical spatial processing employs three parallel encoders operating at distinct patch resolutions.

Input sequences are decomposed into multi-granularity patch representations: $\mathcal{P}^{(i)} = \text{PatchPartition}(\mathcal{V}_{t-k:t}, r_i)$ for $i \in \{1,2,3\}$, where $r_i \in \{4, 8, 16\}$ correspond to fine, medium, and coarse spatial resolutions respectively. Each granularity captures complementary spatial information: fine-scale patches $\mathcal{P}^{(1)} \in \mathbb{R}^{k \times (H/4 \times W/4) \times C}$ preserve detailed texture information for detecting subtle appearance anomalies, medium-scale patches $\mathcal{P}^{(2)} \in \mathbb{R}^{k \times (H/8 \times W/8) \times C}$ balance detail preservation with computational efficiency, and coarse-scale patches $\mathcal{P}^{(3)} \in \mathbb{R}^{k \times (H/16 \times W/16) \times C}$ capture global scene context for understanding large-scale structural changes.

Each spatial scale is processed through Mamba encoders utilizing Multi-Scale Visual State Space Blocks (MS-VSSB), as shown in Fig.~\ref{fig:overall_architecture}(b). The MS-VSSB extends standard VSSB (Fig.~\ref{fig:overall_architecture}(c)) with parallel depth-wise convolutions of varying kernel sizes to capture multi-scale spatial dependencies:
\begin{align}
\mathcal{X}^{(i)} &= \sum_{j \in \{1,3,5\}} \text{DWConv}_{j \times j}(\mathcal{P}^{(i)}), \\
\mathcal{G}^{(i)} &= \text{SpatialEncoder}_i(\mathcal{X}^{(i)} + \mathcal{P}^{(i)}),
\end{align}
where $\mathcal{G}^{(i)} \in \mathbb{R}^{k \times N_i \times D}$ represents the spatial features at scale $i$, and $N_i$ denotes the spatial sequence length at each granularity. Spatial features from different scales are integrated through adaptive importance weighting: $\beta_i = \text{MLP}(\text{GlobalPool}(\mathcal{G}^{(i)}))$ for $i \in \{1,2,3\}$. The final spatial representation is obtained:
\begin{equation}
\mathcal{G}_{\text{spatial}} = \sum_{i=1}^{3} \text{Softmax}([\beta_1, \beta_2, \beta_3])_i \cdot \text{Resize}(\mathcal{G}^{(i)}, N_1),
\end{equation}
where Resize operations standardize different scales to unified spatial dimensions, yielding $\mathcal{G}_{\text{spatial}} \in \mathbb{R}^{k \times N_1 \times D}$.

\subsection{Multi-Scale Temporal Normality Learning}

Temporal normality learning models motion dynamics across varying time horizons to capture anomalous behaviors that occur over different temporal durations. Three temporal windows are constructed to analyze motion patterns at multiple scales: short-term clips $\mathcal{V}_{t-3:t}$ for instantaneous motion changes, medium-term clips $\mathcal{V}_{t-7:t}$ for action-level dynamics, and long-term clips $\mathcal{V}_{t-15:t}$ for behavioral trend analysis. For computational efficiency, temporal motion is represented through frame differencing rather than optical flow computation: $\mathcal{D}_{t-w_j:t-1} = \mathcal{V}_{t-w_j+1:t} - \mathcal{V}_{t-w_j:t-1}$ where $w_j$ denotes the window size for temporal scale $j$.
Each temporal scale employs specialized Temporal Mamba Blocks (TMB) optimized for motion sequence modeling. TMB incorporates temporal positional encoding and scale-specific parameterization:
\begin{align}
\mathcal{A}_j &= -\exp(\boldsymbol{\delta}_j) \odot \boldsymbol{\Lambda}_j, \\
\mathcal{B}_j &= \boldsymbol{\delta}_j \odot \sigma(\text{Linear}(\mathcal{D}_{t-w_j:t-1})), \\
\mathcal{C}_j &= \sigma(\text{Linear}(\mathcal{D}_{t-w_j:t-1})),
\end{align}
where $\boldsymbol{\delta}_j$ represents scale-specific step parameters, $\boldsymbol{\Lambda}_j$ denotes diagonal state matrices, and $\sigma$ is the activation function. Temporal features are computed by integrating motion differences with positional information: $\mathcal{H}^{(j)} = \text{TemporalEncoder}_j(\mathcal{D}_{t-w_j:t-1} + \text{PE}_{t-w_j:t-1})$, 
where PE represents temporal positional encoding. Multi-temporal features $\mathcal{H}_{\text{temporal}}$ are fused through attention-based aggregation that automatically determines the relevance of each temporal scale.

\subsection{Feature Decomposition and Decoding}

The fused spatial-temporal features undergo decomposition to enable specialized processing for appearance and motion reconstruction tasks. The decomposition separates shared representations into task-common and task-specific components through feature disentanglement: $\mathcal{F}_{\text{fused}} = \text{LN}(\mathcal{G}_{\text{spatial}} + \mathcal{H}_{\text{temporal}})$, where learnable gating mechanisms extract common features $\mathcal{F}_{\text{common}} = \text{Sigmoid}(\text{MLP}_{\text{common}}(\mathcal{F}_{\text{fused}})) \odot \mathcal{F}_{\text{fused}}$ and task-specific features $\mathcal{F}_{\text{app}}, \mathcal{F}_{\text{motion}}$ from the residual components. To constrain the representation capability on anomalous samples, memory networks are incorporated following established practices to store prototypical normal patterns. Three memory banks $\{M_c, M_a, M_m\}$ corresponding to common, appearance, and motion features respectively are employed to record normal prototypes through reading and writing operations during training \cite{park2020learning}. The memory mechanism restricts the model's ability to reconstruct anomalous patterns by enforcing consistency with stored normal prototypes \cite{liu2023amp}.
Specialized decoders then process the decomposed features, with the appearance decoder combining common and appearance-specific features to generate frame predictions $\hat{V}_{t+1}$, while the motion decoder fuses common and motion-specific features to reconstruct motion fields $\hat{M}_t$.

\subsection{Loss Function}

Training is guided by a loss function balancing appearance reconstruction, motion modeling, and feature specialization:
$\mathcal{L}_{\text{total}} = \mathcal{L}_{\text{frame}} + \lambda_m \mathcal{L}_{\text{motion}} + \lambda_s \mathcal{L}_{\text{separate}}$, where the frame prediction loss combines pixel-level and gradient consistency:
\begin{equation}
\mathcal{L}_{\text{frame}} = \|\hat{V}_{t+1} - V_{t+1}\|_2^2 + \lambda_g \|\nabla \hat{V}_{t+1} - \nabla V_{t+1}\|_1,
\end{equation}
the motion loss ensures accurate dynamics modeling:
\begin{equation}
\mathcal{L}_{\text{motion}} = \|\hat{M}_t - M_t\|_2^2 + \lambda_{ssim}(1 - \text{SSIM}(\hat{M}_t, M_t)),
\end{equation}
and the separation loss promotes task specialization:
\begin{equation}
    \scalebox{0.88}{$
\mathcal{L}_{\text{separate}} = -\text{CosineSim}(\mathcal{F}_{\text{app}}, \mathcal{F}_{\text{motion}}) + \|\mathcal{F}_{\text{app}} \mathcal{F}_{\text{motion}}^T\|_F^2.
$}
\end{equation}

\subsection{Anomaly Scoring}

During inference, anomaly scores are computed by combining reconstruction errors from both modalities. Frame prediction and motion reconstruction errors are first converted to PSNR values: $\text{PSNR}_{\text{frame}} = 10 \log_{10}\left(\frac{\text{MAX}^2}{\text{MSE}(\hat{V}_{t+1}, V_{t+1})}\right)$ and $\text{PSNR}_{\text{motion}} = 10 \log_{10}\left(\frac{\text{MAX}^2}{\text{MSE}(\hat{M}_t, M_t)}\right)$, where MAX represents the maximum possible pixel value and MSE denotes the mean squared error. The weighted PSNR combination is then computed as $\text{PSNR}_{\text{combined}} = \alpha \cdot \text{PSNR}_{\text{frame}} + (1-\alpha) \cdot \text{PSNR}_{\text{motion}}$, where $\alpha$ balances the contribution of appearance and motion reconstruction quality. Finally, anomaly scores are obtained through min-max normalization across the entire test sequence. 
\section{Experiments}

\subsection{Experiment Preparation}

\noindent \textbf{Datasets.} We evaluate M²S²L on three VAD benchmark datasets: \textit{UCSD Ped2}~\cite{li2013anomaly} contains pedestrian scenes with 16 training and 12 test videos, focusing on anomalies such as bicycles and vehicles in pedestrian areas. \textit{CUHK Avenue}~\cite{lu2013abnormal} comprises 16 training and 21 test videos from university campus scenarios, featuring diverse anomalies including running and wrong-direction movement. \textit{ShanghaiTech}~\cite{luo2017revisit} provides the most challenging evaluation with 330 training and 107 test videos across 13 different scenes. 

\noindent \textbf{Evaluation Metrics.} Following standard evaluation protocols, we adopt frame-level Area Under the ROC Curve (AUC) as the primary detection metric. Besides, we measure Floating Point Operations (FLOPs), model parameters (Params), and inference speed (FPS) on standard hardware configurations.


\noindent \textbf{Implementation Details.} M²S²L processes video clips with temporal window size $k=16$ frames at resolution $256 \times 256$. The multi-scale spatial encoders utilize patch sizes of $\{4, 8, 16\}$ pixels, while temporal encoders operate on windows of $\{4, 8, 16\}$ frames for short, medium, and long-term modeling respectively. Feature dimension $D$ is set to 256 throughout the architecture. Training employs Adam optimizer with initial learning rate $2 \times 10^{-4}$, reduced by factor 0.5 every 20 epochs. Loss function weights are configured as $\lambda_m = 0.5$, $\lambda_s = 0.1$, and $\lambda_g = 0.2$. The PSNR balance parameter $\alpha$ is dataset-specific: 0.6 for UCSD Ped2, 0.4 for CUHK Avenue, and 0.5 for ShanghaiTech.

\subsection{Quantitative AUC and Efficiency Comparison}

Table~\ref{tab:comparison} presents comprehensive comparisons with existing methods, categorized by architectural paradigm and backbone technology. M²S²L achieves competitive detection accuracy of 98.5\%, 92.1\%, and 77.9\% AUC on the three datasets respectively, while maintaining substantial computational advantages. Among single-stream methods, CNN-based approaches like FFP \cite{liu2018future} achieve reasonable efficiency but suffer from limited spatial-temporal modeling capabilities. LSTM-enhanced methods such as ConvLSTM-AE \cite{luo2017remembering} improve temporal modeling but remain constrained by sequential processing. Transformer-based methods \cite{yu2023transanomaly} provide better context modeling but incur quadratic computational complexity. 

Multi-stream approaches demonstrate superior detection performance through specialized appearance and motion processing, but often at significant computational cost. CNN-based multi-stream methods like STD \cite{chang2022video} and AMMC-Net \cite{cai2021appearance} achieve strong results but require substantial computational resources due to their dual-encoder architectures. Our M²S²L, leveraging Mamba backbone with dual-stream architecture, outperforms existing methods while maintaining efficiency: 87\% fewer FLOPs than FFP and 82\% fewer than AMMC-Net \cite{cai2021appearance}. Compared to recent Mamba-based methods, M²S²L surpasses STNMamba \cite{li2024stnmamba} by 1.2\%, 3.1\%, and 3.0\% on the three datasets, validating the effectiveness of multi-scale processing and feature decomposition. The performance improvements are particularly pronounced on complex datasets, where the multi-scale spatial-temporal learning proves crucial for capturing diverse anomaly patterns across different scales and temporal durations. Additionally, M²S²L maintains practical deployment feasibility with 45 FPS inference speed, making it suitable for real-time surveillance applications where both accuracy and efficiency are critical requirements.

\begin{table}[t]
\centering
\caption{Comparison with existing unsupervised methods on the UCSD Ped2 (UP), CHUK Avenue (CA), and ShanghaiTech (ST) datasets.}
\label{tab:comparison}
\resizebox{.5\textwidth}{!}{%
\begin{tabular}{l|l|l|ccc|ccc}
\hline
\multirow{2}{*}{Type} & \multirow{2}{*}{Method} & \multirow{2}{*}{Backbone} & \multicolumn{3}{c|}{AUC (\%)} & \multicolumn{3}{c}{Efficiency} \\
& & & UP & CA & ST & FLOPs (G) & Params (M) & FPS \\
\hline
\multirow{8}{*}{\begin{sideways}Single-Stream\end{sideways}} 
& Conv-AE~\cite{hasan2016learning} & CNN & 90.0 & 70.2 & 60.9 & - & - & - \\
& ConvLSTM-AE~\cite{luo2017remembering} & CNN+LSTM & 88.1 & 77.0 & - & - & - & 10 \\
& sRNN-AE~\cite{luo2017remembering} & LSTM & 92.2 & 83.5 & 69.6 & - & - & 10 \\
& MemAE~\cite{gong2019memorizing} & CNN & 94.1 & 83.3 & 71.2 & 33.0 & \textbf{6.5} & 38 \\
& FFP~\cite{liu2018future} & CNN & 95.4 & 84.9 & 72.8 & 148.1 & 24.2 & 25 \\
& TransAnomaly~\cite{yu2023transanomaly} & Transformer & 96.4 & 87.0 & - & - & - & 18 \\
& AnoPCN~\cite{ye2019anopcn} & CNN & 96.8 & 86.2 & 73.6 & - & - & 10 \\
& MNAD~\cite{park2020learning} & CNN & 97.0 & 88.5 & 70.5 & 46.6 & 15.6 & 65 \\
\hline
\multirow{8}{*}{\begin{sideways}Multi-Stream\end{sideways}}
& STD~\cite{chang2022video} & CNN & 96.7 & 87.1 & 73.7 & 47.9 & 45.1 & 32 \\
& AMMC-Net~\cite{cai2021appearance} & CNN & 96.6 & 86.6 & 73.7 & 169.5 & 25.0 & 18 \\
& MAAM-Net~\cite{wang2023memory} & CNN & 97.7 & 90.9 & 71.3 & - & - & - \\
& CL-Net~\cite{qiu2024video} & CNN & 92.2 & 86.2 & 73.6 & - & - & - \\
& VADMamba~\cite{lyu2025vadmamba} & Mamba & 98.5 & 91.5 & 77.0 & - & 28.1 & \textbf{90} \\
& STNMamba~\cite{li2024stnmamba} & Mamba & 98.0 & 89.0 & 74.9 & \textbf{1.5} & 7.2 & 40 \\
& \textbf{M²S²L (Ours)} & Mamba & \textbf{98.5} & \textbf{92.1} & \textbf{77.9} & 20.1 & 14.9 & 45 \\
\hline
\end{tabular}
}
\vspace{-15pt}
\end{table}

\subsection{Ablation Analysis}

We conduct comprehensive ablation studies to validate each component's contribution in M²S²L. Table~\ref{tab:ablation_components} examines key architectural components through progressive integration. The baseline model (M1) achieves 94.2\%, 88.7\%, and 75.1\% AUC respectively. Multi-scale spatial learning (M2) provides substantial improvements of 3.1\%, 2.4\%, and 2.2\%, demonstrating the critical value of hierarchical spatial feature extraction. Adding multi-scale temporal learning (M3) contributes additional gains of 0.8\%, 0.7\%, and 0.5\%, validating multi-temporal modeling. Feature decomposition (M4) yields consistent gains of 0.4\%, 0.3\%, and 0.1\%, highlighting task-specific separation benefits. The complete M²S²L achieves cumulative improvements of 4.3\%, 3.4\%, and 2.8\% over baseline.

\begin{table}[t]
\centering
\caption{Progressive component ablation study across three benchmark datasets.}
\label{tab:ablation_components}
\begin{tabular}{c|l|ccc}
\hline
Model ID & Configuration &  Ped2 &  Avenue & ShanghaiTech \\
\hline
M1 & Baseline & 94.2 & 88.7 & 75.1 \\
M2 & M1 + MSpatial & 97.3 & 91.1 & 77.3 \\
M3 & M2 + MTemporal & 98.1 & 91.8 & 77.8 \\
M4 & M3 + Decompose & \textbf{98.5} & \textbf{92.1} & \textbf{77.9} \\
\hline
\end{tabular}
\vspace{-15pt}
\end{table}

Table~\ref{tab:ablation_loss} analyzes the impact of different loss function combinations through progressive integration. Using only frame prediction loss (L1) establishes baseline reconstruction capability with 95.8\%, 89.4\%, and 76.3\% AUC. Incorporating motion loss (L2) provides significant improvements of 2.2\%, 2.4\%, and 1.4\% respectively, confirming the substantial value of motion modeling for comprehensive anomaly detection. Finally, adding the separation loss (L3) contributes minor additional improvements of 0.5\%, 0.3\%, and 0.2\% by encouraging task specialization. 

\begin{table}[t]
\centering
\caption{oss function ablation study across three datasets.}
\label{tab:ablation_loss}
\begin{tabular}{c|l|ccc}
\hline
Model & Loss Configuration &  Ped2 &  Avenue & ShanghaiTech \\
\hline
L1 & $\mathcal{L}_{\text{frame}}$ only & 95.8 & 89.4 & 76.3 \\
L2 & L1 + $\mathcal{L}_{\text{motion}}$ & 98.0 & 91.8 & 77.7 \\
L3 & L2 + $\mathcal{L}_{\text{separate}}$ & \textbf{98.5} & \textbf{92.1} & \textbf{77.9} \\
\hline
\end{tabular}
\vspace{-15pt}
\end{table}

\section{Conclusion}

In this paper, we propose a Mamba-based multi-scale spatial-temporal learning framework named  M²S²L for unsupervised video anomaly detection. Compared to existing methods, it includes a dual-stream multi-scale normality learning that separately processes appearance and motion patterns across different spatial granularities and temporal durations, enabling comprehensive anomaly modeling while maintaining computational efficiency. Extensive experiments demonstrate that M²S²L achieves competitive detection performance while requiring significantly lower computational resources with higher inference speed. The ablation studies validate the effectiveness of each proposed component and optimization item, confirming that M²S²L successfully balances detection accuracy with computational feasibility, making it well-suited for real-time video surveillance applications. Future work will explore adaptive scanning mechanisms to further improve the capability in handling real-world surveillance scenarios. 

\bibliographystyle{IEEEtran}
\bibliography{refs.bib}

\begin{thebibliography}{10}
\providecommand{\url}[1]{#1}
\csname url@samestyle\endcsname
\providecommand{\newblock}{\relax}
\providecommand{\bibinfo}[2]{#2}
\providecommand{\BIBentrySTDinterwordspacing}{\spaceskip=0pt\relax}
\providecommand{\BIBentryALTinterwordstretchfactor}{4}
\providecommand{\BIBentryALTinterwordspacing}{\spaceskip=\fontdimen2\font plus
\BIBentryALTinterwordstretchfactor\fontdimen3\font minus \fontdimen4\font\relax}
\providecommand{\BIBforeignlanguage}[2]{{%
\expandafter\ifx\csname l@#1\endcsname\relax
\typeout{** WARNING: IEEEtran.bst: No hyphenation pattern has been}%
\typeout{** loaded for the language `#1'. Using the pattern for}%
\typeout{** the default language instead.}%
\else
\language=\csname l@#1\endcsname
\fi
#2}}
\providecommand{\BIBdecl}{\relax}
\BIBdecl

\bibitem{liu2025networking}
J.~Liu, Y.~Liu, J.~Lin, J.~Li, L.~Cao, P.~Sun, B.~Hu, L.~Song, A.~Boukerche, and V.~C. Leung, ``Networking systems for video anomaly detection: A tutorial and survey,'' \emph{ACM Computing Surveys}, vol.~57, no.~10, pp. 1--37, 2025.

\bibitem{wu2024domain}
T.~Wu, Q.~Chen, D.~Zhao, J.~Wang, and L.~Jiang, ``Domain adaptation of time series via contrastive learning with task-specific consistency,'' \emph{Applied Intelligence}, vol.~54, no.~23, pp. 12\,576--12\,588, Dec. 2024.

\bibitem{liu2025edge}
J.~Liu, Y.~Du, K.~Yang, J.~Wu, Y.~Wang, X.~Hu, Z.~Wang, Y.~Liu, P.~Sun, A.~Boukerche \emph{et~al.}, ``Edge-cloud collaborative computing on distributed intelligence and model optimization: A survey,'' \emph{arXiv preprint arXiv:2505.01821}, 2025.

\bibitem{liu2025domain}
J.~Liu, W.~Zhu, D.~Li, X.~Hu, and L.~Song, ``Domain generalization with semi-supervised learning for people-centric activity recognition,'' \emph{Science China Information Sciences}, vol.~68, no.~1, p. 112103, Jan. 2025.

\bibitem{10646471}
W.~Zhou and Z.~Wang, ``Perceptual depth quality assessment of stereoscopic omnidirectional images,'' \emph{IEEE Transactions on Circuits and Systems for Video Technology}, vol.~34, no.~12, pp. 13\,452--13\,462, 2024.

\bibitem{10409541}
W.~Zhou, R.~Zhang, L.~Li, G.~Yue, J.~Gong, H.~Chen, and H.~Liu, ``Dehazed image quality evaluation: From partial discrepancy to blind perception,'' \emph{IEEE Transactions on Intelligent Vehicles}, vol.~9, no.~2, pp. 3843--3858, 2024.

\bibitem{akdag2024teg}
E.~Akdag, E.~Bondarev, and P.~H. De~With, ``Teg: Temporal-granularity method for anomaly detection with attention in smart city surveillance,'' in \emph{2024 IEEE International Conference on Visual Communications and Image Processing (VCIP)}.\hskip 1em plus 0.5em minus 0.4em\relax IEEE, 2024, pp. 1--5.

\bibitem{liu2023stochastic}
Y.~Liu, D.~Yang, G.~Fang, Y.~Wang, D.~Wei, M.~Zhao, K.~Cheng, J.~Liu, and L.~Song, ``Stochastic video normality network for abnormal event detection in surveillance videos,'' \emph{Knowledge-Based Systems}, vol. 280, p. 110986, Nov. 2023.

\bibitem{zhou2025video}
H.~Zhou, J.~Cai, Y.~Ye, Y.~Feng, C.~Gao, J.~Yu, Z.~Song, and W.~Yang, ``Video anomaly detection with motion and appearance guided patch diffusion model,'' in \emph{Proceedings of the AAAI Conference on Artificial Intelligence}, vol.~39, no.~10, 2025, pp. 10\,761--10\,769.

\bibitem{liu2024generalized}
Y.~Liu, D.~Yang, Y.~Wang, J.~Liu, J.~Liu, A.~Boukerche, P.~Sun, and L.~Song, ``Generalized video anomaly event detection: Systematic taxonomy and comparison of deep models,'' \emph{ACM Computing Surveys}, vol.~56, no.~7, pp. 1--38, 2024.

\bibitem{liu2025anomaly}
Y.~Liu, J.~Liu, C.~Li, R.~Xi, W.~Li, L.~Cao, J.~Wang, L.~T. Yang, J.~Yuan, and W.~Zhou, ``Anomaly detection and generation with diffusion models: A survey,'' \emph{arXiv preprint arXiv:2506.09368}, 2025.

\bibitem{liu2025privacy}
Y.~Liu, S.~Liu, X.~Zhu, J.~Li, H.~Yang, L.~Teng, J.~Guo, Y.~Wang, D.~Yang, and J.~Liu, ``Privacy-preserving video anomaly detection: A survey,'' \emph{IEEE Transactions on Neural Networks and Learning Systems}, pp. 1--22, 2025.

\bibitem{zhao2025rethinking}
M.~Zhao, X.~Zeng, Y.~Liu, J.~Liu, and C.~Pang, ``Rethinking prediction-based video anomaly detection from local--global normality perspective,'' \emph{Expert Systems with Applications}, vol. 262, p. 125581, 2025.

\bibitem{liu2018future}
W.~Liu, W.~Luo, D.~Lian, and S.~Gao, ``Future frame prediction for anomaly detection--a new baseline,'' in \emph{Proceedings of the IEEE Conference on Computer Vision and Pattern Recognition}, 2018, pp. 6536--6545.

\bibitem{liu2023learning}
Y.~Liu, Z.~Xia, M.~Zhao, D.~Wei, Y.~Wang, L.~Siao, B.~Ju, G.~Fang, J.~Liu, and L.~Song, ``Learning causality-inspired representation consistency for video anomaly detection,'' in \emph{Proceedings of the 31st ACM International Conference on Multimedia}, 2023, pp. 203--212.

\bibitem{cheng2024denoising}
K.~Cheng, Y.~Pan, Y.~Liu, X.~Zeng, and R.~Feng, ``Denoising diffusion-augmented hybrid video anomaly detection via reconstructing noised frames,'' in \emph{Proceedings of the Thirty-Third International Joint Conference on Artificial Intelligence}, 2024.

\bibitem{chang2022video}
Y.~Chang, Z.~Tu, W.~Xie, B.~Luo, S.~Zhang, H.~Sui, and J.~Yuan, ``Video anomaly detection with spatio-temporal dissociation,'' \emph{Pattern Recognition}, vol. 122, p. 108213, 2022.

\bibitem{liu2024memory}
Y.~Liu, B.~Ju, D.~Yang, L.~Peng, D.~Li, P.~Sun, C.~Li, H.~Yang, J.~Liu, and L.~Song, ``Memory-enhanced spatial-temporal encoding framework for industrial anomaly detection system,'' \emph{Expert Systems with Applications}, vol. 250, p. 123718, 2024.

\bibitem{rodrigues2021multi}
R.~Rodrigues, N.~Bhargava, R.~Velmurugan, and S.~Chaudhuri, ``Multi-timescale trajectory prediction for abnormal human activity detection,'' \emph{Proceedings of the IEEE/CVF Winter Conference on Applications of Computer Vision}, pp. 2626--2634, 2021.

\bibitem{liu2022appearance}
Y.~Liu, J.~Liu, J.~Lin, M.~Zhao, and L.~Song, ``Appearance-motion united auto-encoder framework for video anomaly detection,'' \emph{IEEE Transactions on Circuits and Systems II: Express Briefs}, vol.~69, no.~5, pp. 2498--2502, 2022.

\bibitem{liu2025crcl}
Y.~Liu, H.~Wang, Z.~Wang, X.~Zhu, J.~Liu, P.~Sun, R.~Tang, J.~Du, V.~Leung, and L.~Song, ``Crcl: Causal representation consistency learning for anomaly detection in surveillance videos,'' \emph{IEEE Transactions on Image Processing}, vol.~34, pp. 2351--2366, 2025.

\bibitem{huang2022multimodal}
C.~Huang, W.~Huang, Q.~Jiang, W.~Wang, J.~Wen, and B.~Zhang, ``Multimodal evidential learning for open-world weakly-supervised video anomaly detection,'' \emph{IEEE Transactions on Multimedia}, vol.~25, pp. 8977--8986, 2022.

\bibitem{gu2023mamba}
A.~Gu and T.~Dao, ``Mamba: Linear-time sequence modeling with selective state spaces,'' \emph{arXiv preprint arXiv:2312.00752}, 2023.

\bibitem{li2024stnmamba}
Z.~Li, M.~Zhao, X.~Yang, Y.~Liu, J.~Sheng, X.~Zeng, T.~Wang, K.~Wu, and Y.-G. Jiang, ``Stnmamba: Mamba-based spatial-temporal normality learning for video anomaly detection,'' \emph{IEEE Transactions on Multimedia}, 2025.

\bibitem{lyu2025vadmamba}
J.~Lyu, M.~Zhao, J.~Hu, X.~Huang, Y.~Chen, and S.~Du, ``Vadmamba: Exploring state space models for fast video anomaly detection,'' \emph{arXiv preprint arXiv:2503.21169}, 2025.

\bibitem{park2020learning}
H.~Park, J.~Noh, and B.~Ham, ``Learning memory-guided normality for anomaly detection,'' \emph{Proceedings of the IEEE/CVF Conference on Computer Vision and Pattern Recognition}, pp. 14\,372--14\,381, 2020.

\bibitem{liu2023amp}
Y.~Liu, J.~Liu, K.~Yang, B.~Ju, S.~Liu, Y.~Wang, D.~Yang, P.~Sun, and L.~Song, ``Amp-net: Appearance-motion prototype network assisted automatic video anomaly detection system,'' \emph{IEEE Transactions on Industrial Informatics}, vol.~20, no.~2, pp. 2843--2855, 2023.

\bibitem{li2013anomaly}
W.~Li, V.~Mahadevan, and N.~Vasconcelos, ``Anomaly detection and localization in crowded scenes,'' in \emph{IEEE Transactions on Pattern Analysis and Machine Intelligence}, vol.~36, no.~1, 2014, pp. 18--32.

\bibitem{lu2013abnormal}
C.~Lu, J.~Shi, and J.~Jia, ``Abnormal event detection at 150 fps in matlab,'' in \emph{Proceedings of the IEEE International Conference on Computer Vision}, 2013, pp. 2720--2727.

\bibitem{luo2017revisit}
W.~Luo, W.~Liu, and S.~Gao, ``A revisit of sparse coding based anomaly detection in stacked rnn framework,'' \emph{Proceedings of the IEEE International Conference on Computer Vision}, pp. 341--349, 2017.

\bibitem{luo2017remembering}
W.~Luo and S.~Gao, ``Remembering history with convolutional lstm for anomaly detection,'' \emph{Proceedings of the IEEE International Conference on Multimedia and Expo}, pp. 439--444, 2017.

\bibitem{yu2023transanomaly}
G.~Yu, S.~Wang, Z.~Cai, E.~Zhu, C.~Xu, J.~Yin, and M.~Kloft, ``Transanomaly: Video anomaly detection using video vision transformer,'' \emph{IEEE Access}, vol.~9, pp. 123\,977--123\,986, 2021.

\bibitem{cai2021appearance}
R.~Cai, H.~Zhang, W.~Liu, S.~Gao, and Z.~Hao, ``Appearance-motion memory consistency network for video anomaly detection,'' \emph{Proceedings of the AAAI Conference on Artificial Intelligence}, vol.~35, no.~2, pp. 938--946, 2021.

\bibitem{hasan2016learning}
M.~Hasan, J.~Choi, J.~Neumann, A.~K. Roy-Chowdhury, and L.~S. Davis, ``Learning temporal regularity in video sequences,'' in \emph{Proceedings of the IEEE Conference on Computer Vision and Pattern Recognition}, 2016, pp. 733--742.

\bibitem{gong2019memorizing}
D.~Gong, L.~Liu, V.~Le, B.~Saha, M.~R. Mansour, S.~Venkatesh, and A.~v.~d. Hengel, ``Memorizing normality to detect anomaly: Memory-augmented deep autoencoder for unsupervised anomaly detection,'' \emph{Proceedings of the IEEE/CVF International Conference on Computer Vision}, pp. 1705--1714, 2019.

\bibitem{ye2019anopcn}
M.~Ye, X.~Peng, W.~Gan, W.~Wu, and Y.~Qiao, ``Anopcn: Video anomaly detection via deep predictive coding network,'' in \emph{Proceedings of the 27th ACM International Conference on Multimedia}, 2019, pp. 1805--1813.

\bibitem{wang2023memory}
L.~Wang, J.~Tian, S.~Zhou, H.~Shi, and G.~Hua, ``Memory-augmented appearance-motion network for video anomaly detection,'' \emph{Pattern Recognition}, vol. 138, p. 109335, 2023.

\bibitem{qiu2024video}
S.~Qiu, J.~Ye, J.~Zhao, L.~He, L.~Liu, X.~Huang \emph{et~al.}, ``Video anomaly detection guided by clustering learning,'' \emph{Pattern Recognition}, vol. 153, p. 110550, 2024.

\end{thebibliography}

\end{document}